\documentclass[reprint,showpacs,amsmath,amssymb,aps,superscriptaddress,floatfix,longbibliography,nofootinbib,twocolumn]{revtex4-2}
\usepackage{amsthm}
\usepackage{amsmath, mathtools}
\usepackage[utf8]{inputenc}	
\usepackage[english]{babel}
\usepackage{amsmath}
\usepackage{graphicx}		
\usepackage{natbib}
\usepackage{textcomp}
\usepackage{gensymb}
\usepackage{placeins}
\usepackage[usenames,dvipsnames,svgnames,table]{xcolor}
\usepackage[hidelinks,colorlinks=false,urlcolor=Cerulean,citecolor=black]{hyperref}
\usepackage{siunitx}
\usepackage{mathrsfs}
\usepackage{multirow}
\usepackage{bm}
\usepackage{xfrac}
\usepackage{comment}
\usepackage{booktabs}
\usepackage[normalem]{ulem}
\usepackage{float}
\renewcommand{\i}{\mathrm{i}}

\DeclareMathOperator{\R}{\mathbb{R}}
\DeclareMathOperator{\C}{\mathbb{C}}

\DeclareMathOperator\Arg{Arg}
\renewcommand{\i}{\mathrm{i}}

\begin{document}

\title{An explicit operator explains end-to-end computation in the modern neural \\ networks used for sequence and language modeling}

\author{Anif N. Shikder}
\thanks{These authors contributed equally to this work.}
\affiliation{Department of Mathematics, Western University, London ON, Canada}
\affiliation{Fields Lab for Network Computation, Fields Institute, Toronto ON, Canada}

\author{Ramit Dey}
\thanks{These authors contributed equally to this work.}
\affiliation{Department of Mathematics, Western University, London ON, Canada}
\affiliation{Fields Lab for Network Computation, Fields Institute, Toronto ON, Canada}

\author{Sayantan Auddy}
\affiliation{Department of Mathematics, Western University, London ON, Canada}
\affiliation{Fields Lab for Network Computation, Fields Institute, Toronto ON, Canada}

\author{Luisa Liboni}
\affiliation{King's University College at Western University, London ON, Canada}
\affiliation{Fields Lab for Network Computation, Fields Institute, Toronto ON, Canada}

\author{Alexandra Busch}
\affiliation{Department of Mathematics, Western University, London ON, Canada}
\affiliation{Fields Lab for Network Computation, Fields Institute, Toronto ON, Canada}

\author{Arthur Powanwe}
\affiliation{Department of Mathematics, Western University, London ON, Canada}
\affiliation{Fields Lab for Network Computation, Fields Institute, Toronto ON, Canada}

\author{Ján Mináč}
\affiliation{Department of Mathematics, Western University, London ON, Canada}
\affiliation{Fields Lab for Network Computation, Fields Institute, Toronto ON, Canada}

\author{Roberto C. Budzinski}
\affiliation{Department of Neuroscience, University of Lethbridge, Lethbridge AB, Canada}
\affiliation{Fields Lab for Network Computation, Fields Institute, Toronto ON, Canada}

\author{Lyle E. Muller}
\thanks{lmuller2@uwo.ca}
\affiliation{Department of Mathematics, Western University, London ON, Canada}
\affiliation{Fields Lab for Network Computation, Fields Institute, Toronto ON, Canada}

\begin{abstract}
We establish a mathematical correspondence between state space models, a state-of-the-art architecture for capturing long-range dependencies in data, and an exactly solvable nonlinear oscillator network. As a specific example of this general correspondence, we analyze the diagonal linear time-invariant implementation of the Structured State Space Sequence model (S4). The correspondence embeds S4D, a specific implementation of S4, into a ring network topology, in which recent inputs are encoded as waves of activity traveling over the one-dimensional spatial layout of the network. We then derive an exact operator expression for the full forward pass of S4D, yielding an analytical characterization of its complete input–output map. This expression reveals that the nonlinear decoder in the system induces interactions between these information-carrying waves that enable classifying real-world sequences. These results generalize across modern SSM architectures, and show that they admit an exact mathematical description with a clear physical interpretation. These insights enable a new level of interpretability for these systems in terms of nonlinear oscillator networks.
\end{abstract}

\maketitle


Transformers are a central machine learning architecture for language and sequence tasks \cite{vaswani2017attention}. Transformers capture long-range dependencies in language through a mechanism called ``matrix attention'' \cite{bahdanau2014neural,muller2024transformers}. However, because this mechanism requires computing pairwise interactions between all tokens in a sequence -- for example, all words in a sentence -- it introduces a computational bottleneck when handling long-range interactions across inputs \cite{tay2020efficient}. This results in quadratic scaling of both memory and computation with respect to sequence length. This quadratic complexity is, in turn, a key factor for resource consumption in these models as they scale to larger problems and longer context lengths \cite{child2019generating,katharopoulos2020transformers}.

State-space models (SSMs) have recently emerged as an alternative architecture for sequence modeling, addressing the central computational inefficiency of transformers \cite{DBLP:journals/corr/abs-2111-00396, gu2022parameterization, gu2023mamba, smith2022simplified,orvieto2023resurrecting}. Unlike transformers, SSMs achieve linear scaling in sequence length by encoding temporal dependencies in patterns of activity across a network \cite{DBLP:journals/corr/abs-2111-00396}. Because connections between nodes cause each input to reverberate within the network, an SSM can build up a complex dynamical state that contains information about the history of inputs to the system. SSM architectures have demonstrated that this dynamics-based computation can be both efficient and highly parallelizable during training \cite{DBLP:journals/corr/abs-2111-00396}. SSMs now match or exceed transformers on sequence modeling benchmarks critical to language tasks \cite{smith2022simplified}, while also dramatically reducing computational requirements for long sequences \cite{DBLP:journals/corr/abs-2111-00396,smith2022simplified}.

In addition to increased efficiency, SSMs offer potentially substantial advantages in interpretability over transformers:~unlike the attention mechanism in transformers, the dynamics within an SSM are linear and therefore mathematically solvable. This opens the possibility of understanding SSMs through mathematical analyses of their internal network dynamics, offering a level of explainability beyond {\it post-hoc} approaches for interpreting transformer outputs \cite{elhage2021mathematical}, which observe model behavior empirically and case-by-case, rather than deriving it from the model's mathematical structure from first principles. Linear dynamics alone are not sufficient for sequence processing, however: without an additional nonlinear decoder, SSM architectures cannot capture the range of input-output mappings needed to perform sequence processing tasks \cite{NEURIPS2023_ea8608c6,NEURIPS2024_e6231c5f}. Because of the fundamental difficulty in analyzing these interleaved linear and nonlinear operations, end-to-end explainability of SSMs has not yet been achieved. Establishing this link would represent a substantial advance in understanding how an SSM generates output, potentially leading to future large language models whose outputs can be precisely understood and controlled.

We now present a theoretical framework in which the complete linear and nonlinear cascade of an SSM can be understood exactly, revealing the mechanism of computation in these systems. This framework leverages an exact solution for the dynamics of a nonlinear oscillator network \cite{muller2021algebraic} and an exact operator description of computation in this network \cite{budzinski2024exact}. We first establish a formal mathematical correspondence between this oscillator network and diagonal SSMs -- currently the most widely deployed SSM architectures for sequence processing. We then derive an end-to-end mathematical description of computation in S4D, a specific implementation of the Structured State Space Sequence model (S4) \cite{gupta2022diagonal,gu2022parameterization}. This expression reveals that different input sequences excite specific waves traveling over the network, and the nonlinear readout induces interactions between waves that are critical for classifying real-world inputs. This framework thus provides mechanistic insight into computation in SSMs, while also suggesting principled ways to design future architectures.


\section*{Mathematical background}

Consider a network of $N$ nodes arranged on a one-dimensional ring. The network evolves in time according to:
\begin{equation}  
\begin{split} \label{psi}
    \dot{\psi_i} = \omega + \kappa \sum_{j=1}^{N} a_{ij} \Big( \sin\left( \psi_j - \psi_i - \phi_{ij} \right) \\ - \i\cos\left( \psi_j - \psi_i - \phi_{ij} \right) \Big)\,,
\end{split}
\end{equation}
where $\psi_i(t) \in \C$ is the state of node $i$ at time $t$, $\omega \in \R$ is the intrinsic oscillation frequency, the matrix element $a_{ij} \in \R$ represents the connection from node $j$ to $i$, $\kappa \in \R$ homogeneously scales the strength of connections in the network, $\phi_{ij} \in \R$ is a phase lag in the connection, and $\i = \sqrt{-1}$ is the imaginary unit (note that $i$ is an index).

We have recently shown that Eq.~\eqref{psi} admits an exact solution \cite{muller2021algebraic}. To see this, change Eq.~\eqref{psi} to a rotating coordinate frame (in which $\omega$ can be set to zero w.l.o.g.~-- see, for example, \cite{strogatz1988collective}), multiply Eq.~\eqref{psi} by $\i$, and apply standard identities to obtain:
\begin{equation}
    \i\dot{\psi_i} = \kappa e^{-\i\psi_i} \sum_{j=1}^{N} a_{ij}e^{-\i\phi_{ij}} e^{\i\psi_j} \,.
\end{equation}
This system of equations is nonlinear in $\psi_i$; however, by applying the nonlinear coordinate transformation $x_i = e^{\i\psi_i}$ and noticing that $\frac{dx_i}{dt} = \i e^{\i\psi_i}\frac{d\psi_i}{dt}$, we obtain an exact solution in the new variable $x$, expressed in matrix form as:
\begin{equation} \label{exact_sol}
    \bm{x}(t) = e^{\bm{K}t}\bm{x}(0)\,,
\end{equation}
where $\bm{x}(t) \in \C^N$ is the solution for the state vector of the network at time $t$, $e$ is the matrix exponential, the matrix $\bm{K}$ has elements $k_{ij} = \kappa e^{-\i\phi_{ij}} a_{ij}$, and is a composite matrix containing information about the coupling strength $\kappa$, network connectivity $a_{ij}$ and phase lag $\phi_{ij}$, and $\bm{x}(0) \in \C^N$ are the initial conditions. Note that vectors and matrices will be typeset in bold throughout. Note, as well, the exact solution can be expressed in the non-rotating coordinate frame by multiplying the RHS of Eq.\,\eqref{exact_sol} by an additional term $e^{\i\omega t}$. 

In terms of the $\psi$-coordinate, the exact solution is then given by
\begin{equation} \label{exact_sol_psi}
    \bm{\psi}(t)=\Arg\!\left(e^{\bm{K}t}e^{\i \bm{\psi}(0)}\right)-\i\log\!\big|e^{\bm{K}t}e^{\i\bm{\psi}(0)}\big|
\end{equation}
where $\bm{\psi}(0) \in \C^N$ are the initial conditions in terms of the $\psi$-coordinate, and scalar functions (here $\mathrm{Arg}$, $\log$, and modulus) are taken elementwise when applied to a vector. 

Finally, we can utilize the exact solution Eq.~\eqref{exact_sol} to propagate the recurrent dynamics in terms of the $x$-coordinate in discrete steps of time $\tau k$:
\begin{equation}
    \bm{x}_k = e^{\bm{K}\tau k}\bm{x}_{k-1}\,.
    \label{eq:solution_discrete}
\end{equation}
The matrix exponential serves as an exact propagator to bring the continuous-time dynamics from time point $\tau(k-1)$ to $\tau k$. We can then use the exact solution in Eq.~\eqref{eq:solution_discrete} to define a dynamics operator $\mathcal{D}_\tau = e^{\bm{K} \tau}$ that propagates the state of the network forward in time. The operator $\mathcal{D}_\tau$ is applied at discrete time $k-1$ to evolve the state of the network to discrete time $k$. In terms of this operator, we can write Eq.~\eqref{eq:solution_discrete} as $\bm{x}_k = \mathcal{D}_\tau \bm{x}_{k-1}$.

\begin{figure}[b!]
    \centering
    \includegraphics[width=\columnwidth]{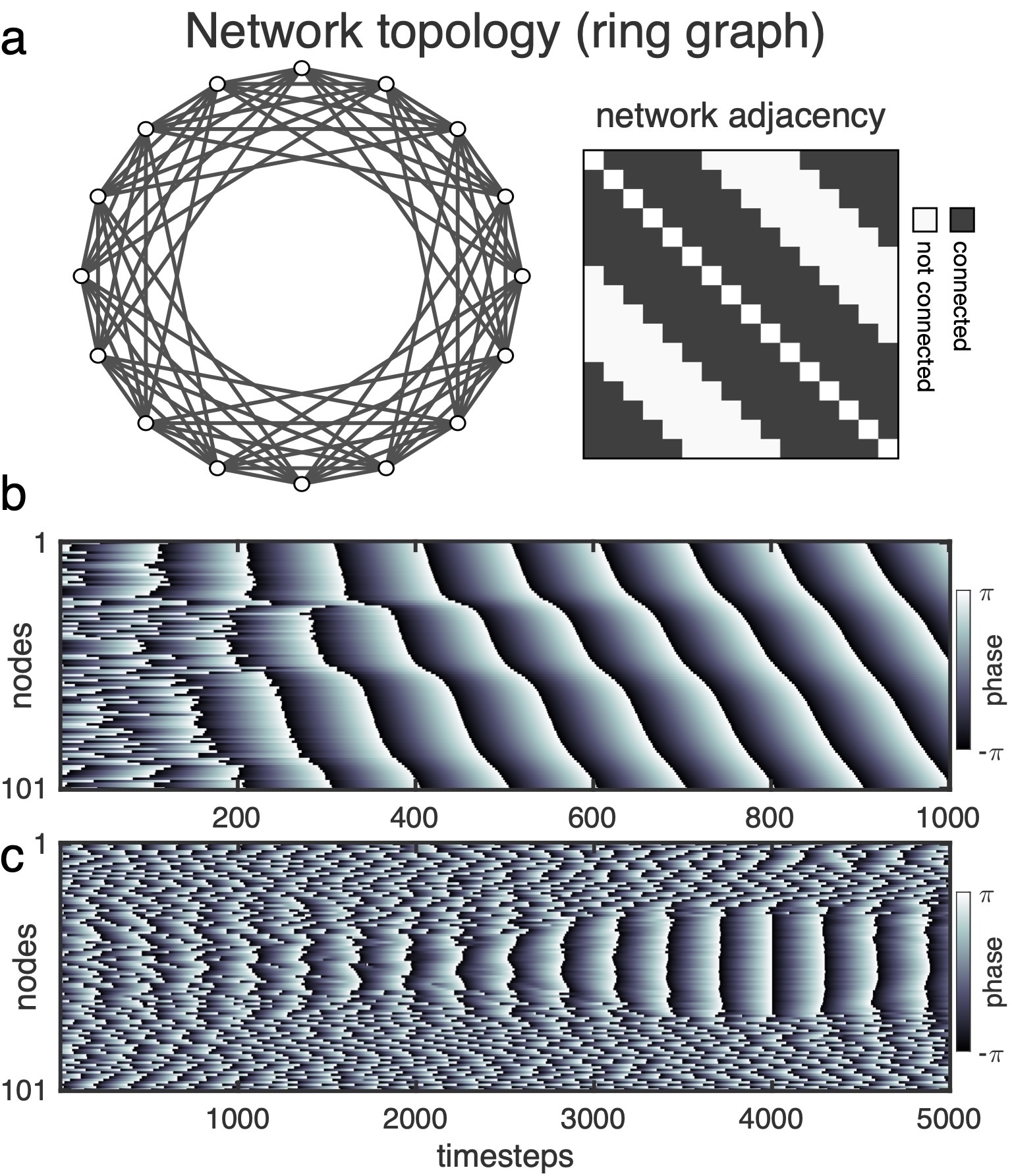}
    \caption{\textbf{The nonlinear oscillator network produces rich spatiotemporal dynamics across its ring topology.} \textbf{(a)} The network is composed of $N$ nodes arranged on a one-dimensional ring (left), resulting in a network adjacency matrix with connections between nodes in a neighborhood of $n$ steps on the ring with boundaries conditions (right). \textbf{(b)} For specific combinations of network connectivity and phase-lags, traveling waves emerge in the phase dynamics ($\Arg(x_j(t))\,\forall\,j$), here plotted in color-code as a function of node position and time. \textbf{(c)} Different spatiotemporal patterns are possible: a localized pocket of synchronization that co-exist with asynchronous behavior, known as a ``chimera'' state \cite{abrams2004chimera,budzinski2024exact}, emerges in the phase dynamics of the network with homogeneous phase lag.}
    \label{fig:cvnn_dynamics}
\end{figure}
Intuition from the theory of coupled oscillators becomes useful at this point: across a range of phase lags, the network exhibits long transients, rich spatiotemporal dynamics, and signatures of spatiotemporal chaos (Fig.~\ref{fig:cvnn_dynamics}). The rich spatiotemporal dynamics in the phase of this system (Fig.~\ref{fig:cvnn_dynamics}b,c) are in fact the result of complex-linear dynamics viewed through the lens of a nonlinear readout, since the phase $\Arg(x_j(t))\,\forall\,j$ is a nonlinear function applied to each element of the state $\bm{x}(t)$. This system, with complex-linear dynamics and a nonlinear readout, has recently been found to allow a range of computations -- from logic gates, short-term memory, and message encryption \cite{budzinski2024exact} to simple computer vision tasks like image segmentation \cite{liboni2025image}, while also being exactly mathematically interpretable. Finally, an input to this network is straightforward to incorporate into this framework: 
\begin{equation}
    \bm{x}_k = \mathcal{D}_\tau \bm{x}_{k-1} + \bm{\bar{B}}\bm{u}_k
    \label{ko20input}
\end{equation}
where $\bm{\bar{B}} = (\Delta \bm{K})^{-1} (e^{\bm{K}\Delta}-I)\Delta\bm{B}$ is the discrete-time input projection to the system and $\bm{u}_k$ is an input vector (e.g.\,following an encoder block) at time point $k$.

The operator $\mathcal{D}_\tau$ can be diagonalized numerically, in general. In addition, many connection patterns in $\bm{K}$ admit fully analytical forms for diagonalizing $\mathcal{D}_\tau$. For example, if the strength of connections from node $i$ to other nodes $j$ falls off with distance (e.g.~with a power law), and the same connection rule is applied to each node $i$, then the adjacency matrix defining this network will be circulant. A circulant matrix is constructed by cyclically permuting a generating vector $\bm{c}$ $N$ times, resulting in an $N \times N$ matrix. The Circulant Diagonalization Theorem (CDT) states that all circulants are diagonalized by the same unitary matrix $\bm{F}$, with entries $f_{ij}$ \cite{davis1979circulant}:
\begin{equation}
    f_{ij} = \frac{1}{\sqrt{N}} \exp\left[ - \frac{2\pi \i}{N} (i-1)(j-1) \right]\,,
    \label{eq:eigenbasis_f}
\end{equation}
and there is a similar closed-form expression for the eigenvalues in terms of the generating vector $\bm{c}$ (see Appendix~\ref{secA1}). Applying the change of basis $\bm{F}^\dagger$ to Eq.\,\eqref{ko20input} then results in:
\begin{equation}
    \tilde{\bm{x}}_k = e^{\bm{\Lambda}\tau} \tilde{\bm{x}}_{k-1} + \bm{\tilde{B}}\bm{u}_k,
    \label{eq:solution_basis}
\end{equation}
where $\tilde{\bm{x}}_k = \bm{F}^\dagger \bm{x}_k$, $\bm{\Lambda}$ is a diagonal matrix containing the eigenvalues $\lambda_s$ of $\bm{K}$ (see Appendix~\ref{secA1}), and $\bm{\tilde{B}}=\bm{F}^\dagger\bm{\bar{B}}$. As demonstrated in the next section, Eq.\,\eqref{eq:solution_basis} allows revealing a precise correspondence with a family of linear time-invariant SSMs used in sequence processing.

\section*{Results}

\subsection*{Correspondence with diagonal SSMs}

Several implementations of S4 achieve high accuracy on benchmarks for sequence processing \cite{DBLP:journals/corr/abs-2111-00396,gupta2022diagonal,gu2022parameterization} such as the long-range arena (LRA) \cite{tay2021long}. We will focus on analyzing S4D-Lin \cite{gu2022parameterization}, which is a diagonalized version of S4 that is most commonly used in implementation. We note that the mathematical analysis introduced here generalizes across all diagonal SSMs (see Appendix~\ref{secA2}). In addition, because the recurrence matrix for nearly all SSMs is diagonalizable, this analysis applies quite generally across SSM architectures. 

The linear recurrence in S4D-Lin evolves in discrete time according to the equation:
\begin{equation}
    \bm{x}_k = e^{\bm{D}\tau}\bm{x}_{k-1} + \bm{\bar{B}}\bm{u}_k\label{s4dlin}
\end{equation}
where $\bm{D}$ is a diagonal matrix with entries $d_{jj}$:
\begin{equation}
    d_{jj} = -\frac{1}{2} + \i\pi j\,.
    \label{eq:_eigenvalue_d}
\end{equation}
Comparing Eq.\,\eqref{s4dlin} with Eq.\,\eqref{eq:solution_basis} reveals the precise correspondence between the exactly solvable nonlinear oscillator network and S4D-Lin:~both systems evolve by a diagonal operator acting on the state vector $\bm{x}_{k-1} \in \mathbb{C}^N$, with externally added input $\bm{u}_k$ linearly projected through the operator $\bm{\bar{B}}$ at each discrete time step. The two systems become fully equivalent by matching the eigenvalue matrices -- $\bm{\Lambda}$ in Eq.\,\eqref{eq:solution_basis} and $\bm{D}$ in Eq.\,\eqref{s4dlin} -- and the input projection matrices.\footnote{Note that $\tau$ used here is equivalent to $\Delta t$ in the work introducing S4D-Lin \cite{gu2022parameterization}.} Importantly, the nonlinear coordinate transformation $x_i = e^{\i\psi_i}$ that renders the oscillator network exactly solvable is also what establishes the mathematical correspondence with S4D-Lin. The central distinction between the systems, then, is that the oscillator network dynamics unfold in a precisely defined eigenbasis given by the unitary operator $\bm{F}$, whereas S4D-Lin leaves this eigenvector basis unspecified. This oscillator correspondence thus provides both an explicit eigenbasis in which the SSM dynamics evolve and, as demonstrated in the following section, a ring topology in which the SSM's latent state can be understood in terms of traveling waves. This theoretical foundation will, in turn, allow an explicit characterization of computation in these systems during sequence classification.

\subsection*{Traveling waves in S4D's recurrent layer \\ distinguish simple inputs}

\begin{figure*}[thb]    
    \centering
    \includegraphics[width=\linewidth]{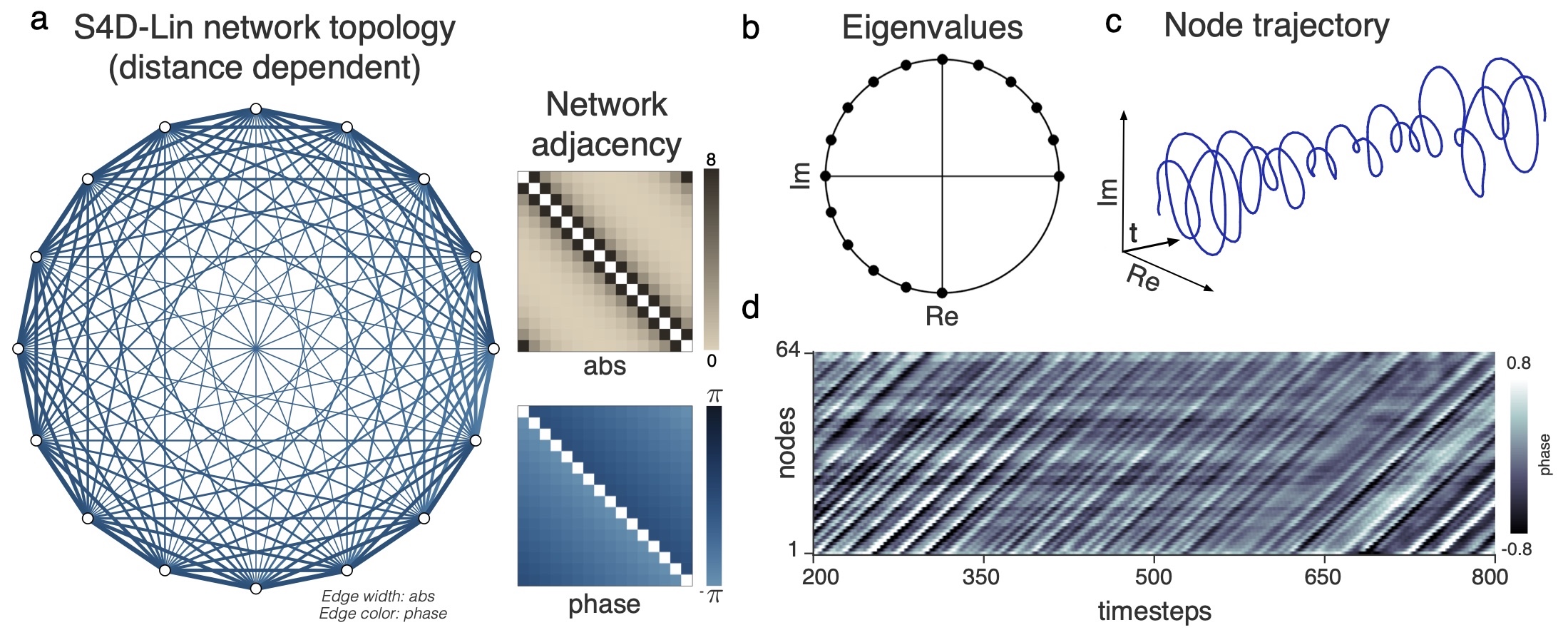}
    \caption{\textbf{The oscillator network correspondence reveals S4D generates traveling waves in its dynamical state.} \textbf{(a)} The mathematical correspondence embeds S4D into a ring topology with a specific structure of network connections, where connection strengths weaken with distance between nodes and phase delays shift systematically. The network adjacency matrix is depicted at right, with magnitude (top) and phase (bottom) of each element plotted in color code. \textbf{(b)} Eigenvalue spectrum of the S4D state-transition operator (i.e. $\mathcal{D}_{\tau}$) in the complex plane. As details in Eq.\,\eqref{eq:_eigenvalue_d}, all modes share a common decay rate (real part), while the imaginary parts increase approximately linearly with mode index, forming a harmonic ladder of temporal frequencies. \textbf{(c)} Representative node trajectory in the complex plane, illustrating oscillatory dynamics with a slowly decaying amplitude. \textbf{(d)} Spatiotemporal evolution of node phases across time. Diagonal phase bands indicate coherent traveling waves propagating around the ring, with velocity determined by the eigenfrequencies of the dynamics operator.}
    \label{fig:dynamics}
\end{figure*}
We can now use the framework introduced in the previous sections to analyze the dynamics in the recurrent layer of S4D. Specifically, using the explicit basis $\bm{F}$ from the nonlinear oscillator context, we can now write the state $\bm{x}_k$ as a superposition of the eigenmodes $\bm{f}_j$:
\begin{equation} \label{x_mu}
\bm{x}_k = \sum_{j=1}^{N} \mu_j(k)\, \bm{f}_j\,.
\end{equation}
Here, $\mu_j(k) = \langle \bm{f}_j | \bm{x}_k \rangle$ is the complex amplitude of the $j$th eigenvector in the basis, which represents a traveling wave in the node dynamics at a specific spatial frequency (cf.\,Eq.\,\ref{eq:eigenbasis_f}). Further, we can use the eigenvalues from S4D (Eq.\,\ref{eq:_eigenvalue_d}) to understand the complex adjacency matrix $\bm{K}$ in the equivalent oscillator network, via the expression $\bm{K} = \bm{F} \bm{D} \bm{F}^\dagger$ (Fig.\,\ref{fig:dynamics}a). When exponentiated (in $e^{\bm{D}\tau}$, Eq.\,\ref{s4dlin}), the eigenvalues $d_{jj}$ fall on the unit circle in the complex plane (Fig.\,\ref{fig:dynamics}b). This matrix and its eigenstructure determines the interactions between nodes, and the spatio-temporal structure of the dynamics that the network will produce \cite{budzinski2022geometry,budzinski2023analytical}.

Without input, and with random initial conditions, the recurrent layer generates transient, fluctuating dynamics at each node (Fig.\,\ref{fig:dynamics}c). These fluctuations in activity at single nodes form waves traveling in a single direction around the network's ring topology (Fig.\,\ref{fig:dynamics}d). Waves traveling over a topographically organized neural network are known to serve useful computational roles:~by embedding a continuous spatiotemporal structure across the network, they can provide a way to store information about the recent past \cite{muller2018cortical,benigno2023waves,keller2024traveling}. This property has been observed with physical experiments in fluid mechanics \cite{perrard2018transition} and in trained recurrent neural networks \cite{keller2023neural}. It has previously been recognized that this property can be a useful way to store long-term dependencies directly in a network's activity structure \cite{muller2024transformers,keller2026spatiotemporal}, but has not previously been expressed in a direct mathematical form. 

We can now show that, when driven by input, S4D indeed stores information about the recent past through traveling waves in its recurrent layer. Since $\mathcal{D}_\tau$ acts diagonally in the $\bm{F}-$basis, each amplitude evolves independently, yielding the simple update rule
\begin{equation}
\mu_i(k) = \lambda_{D,i}\, \mu_i(k-1) + \bm{\tilde{b}}_i^{T}\, \bm{u}_k ,
\label{eq:basic_mu_update}
\end{equation}
where $\lambda_{D,i}$ is the $i$-th eigenvalue of $\mathcal{D}_\tau$ and $\bm{\tilde{b}}_i = [(\bm{F}^{\dagger} \bm{B})_i]^{T}$ describes how strongly the input at time $k$ excites wave $i$. 

\begin{figure*}[thb]    
    \centering
    \includegraphics[width=.95\linewidth]{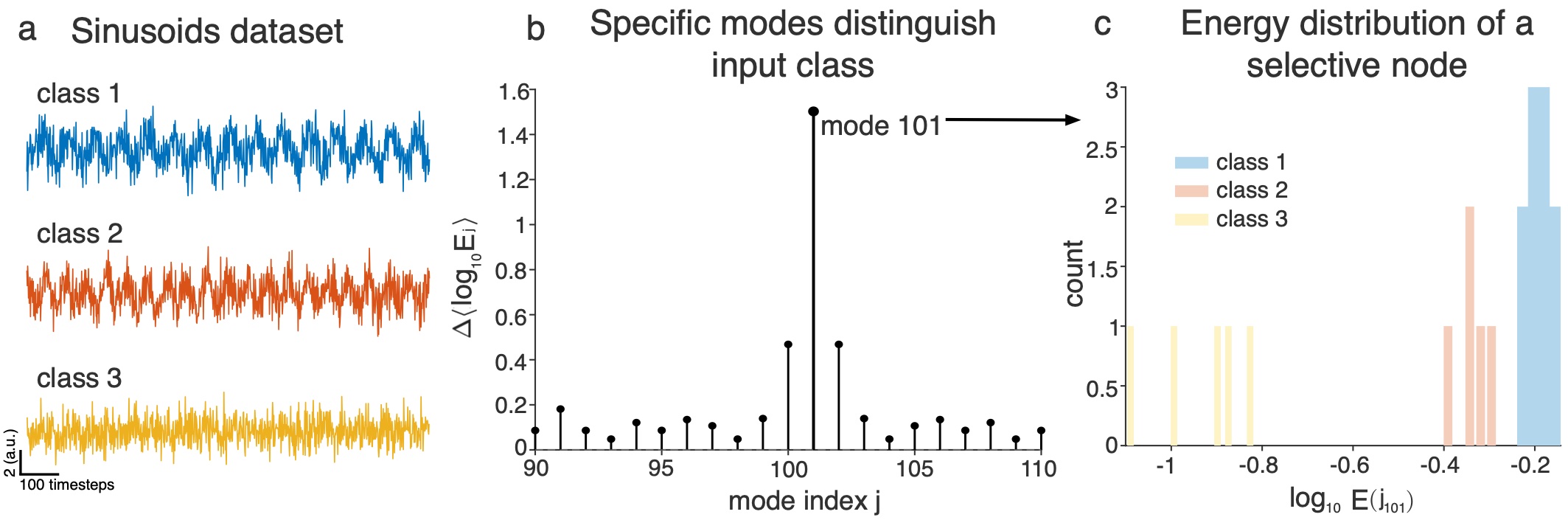}
    \caption{\textbf{Traveling waves distinguish different inputs in a simple dataset.} \textbf{(a)} Representative samples of timeseries data from each of the three classes. Class~1 contains 15~Hz sinusoids embedded in noise; Class~2 contains 20~Hz sinusoids embedded in noise; and Class~3 consists of signals with noise only. \textbf{(b)} We calculate the difference in modal energy $E_j = \sum_k |\mu_j(k)|^2$ across inputs, which is the summed difference between class 1 and 2, 1 and 3, and 2 and 3. Each mode corresponds to an eigenfrequency of the state-transition operator $D_\tau$. Only a small subset of resonant modes shows pronounced differences. The stem plot shows that mode $101$, whose eigenphase lies closest to the 15--20~Hz temporal frequency band, separates input classes most effectively. \textbf{(c)} Single-trial distributions of the resonant modal energy $\log_{10} E(j_{101})$, shown as histograms with counts. This single resonant mode exhibits clear class separation: the 15~Hz and 20~Hz classes concentrate at distinct energy levels, while the noise-only class occupies a lower-energy regime. Together, these results demonstrate that discrimination arises from a small subset of resonant modes rather than from global spectral differences.}
    \label{fig:modal_energy}
\end{figure*}
Equation \eqref{eq:basic_mu_update} shows that the input signal is encoded by waves traveling across the recurrent layer, with temporal frequency determined by $\operatorname{Im}(\lambda_{D,j})$ and decay controlled by $\operatorname{Re}(\lambda_{D,j})$. To demonstrate how different signals are encoded by these waves, we trained S4D to classify a synthetic dataset composed of three different signal classes embedded in high noise (Fig.\,\ref{fig:modal_energy}a). The first class contained ten sinusoids with fixed frequency ($15\text{~Hz}$), random amplitude, and random phase embedded in strong Gaussian noise ($-0.5$ to $-1~\text{dB}$ signal-to-noise ratio). The second class consisted of ten $20\text{~Hz}$ sinusoids, again with random amplitude and phase, and embedded in Gaussian noise with the same strength. The third class contained no sinusoids and only Gaussian noise. All signals were sampled for 900 time steps, and there were 300 independent realizations for each class in the dataset tested here. Following the analytical expression Eq.\,\eqref{eq:basic_mu_update}, when the network is driven by these inputs, the recurrent layer generates waves corresponding to specific modes $\mu_j(k)$ (Fig.\,\ref{fig:modal_energy}b). Summing the square of each mode $(\mu_j(k))^2$ over time shows that power at specific modes distinguishes the different input classes across trials (Fig.\,\ref{fig:modal_energy}c). 

This result demonstrates that S4D generates traveling waves in its recurrent layer that distinguish simple inputs, allowing successful classification at the system's readout. As discussed above, the sinusoidal inputs that are varying in time become translated into waves traveling across the network. These spatial modes allow a relatively small $5\text{~Hz}$ difference in temporal frequency between class 1 and class 2 to build up over time, until the network produces very different traveling wave patterns in the recurrent layer. 

These wave dynamics in the recurrent layer of the SSM are then processed through a series of linear and nonlinear operations. While only the linear recurrent dynamics in S4D-Lin are needed to distinguish inputs in this relatively simple dataset, more complex, real-world inputs require the entire cascade of linear and nonlinear steps. In general, the forward pass of the network -- i.e., how a specific input leads to a specific classification -- is very difficult to analyze mathematically when the interactions between linear and nonlinear steps become critical to the network's performance. In the next section, however, we will introduce a mathematical approach, using the framework introduced above, that allows understanding precisely how S4D classifies inputs in a real-world sequence processing task.

\subsection*{An explicit operator for computation in S4}

Our goal is to write a mathematical expression that exactly characterizes a single computation in S4, from receiving an input to generating an output. To do this, we will utilize the connection between S4D and the oscillator networks defined above. 

This expression should take the following operator form:
\begin{equation}
    \bm{O}
    =
    \frac{1}{T}\,
    \bm{W}\sum_{k=1}^T \Lambda\bigg(\Re\big(\bm{C}\,[\mathcal{D}_\tau \bm{x}_k + \bar{\bm{B}} \bm{u}_k]\big)\bigg)\,,\label{eq:operator}
\end{equation}
relating the input $\bm{u}_k\in\mathbb{R}^{d_{\mathrm{model}}}$ to the output $\bm{O}\in\mathbb{R}^{N_{\mathrm{c}}}$ (where $N_{\mathrm{c}}$ is the number of output classes) via the linear transformation $\bm{C}\in\mathbb{R}^{d_{\mathrm{model}}\times N}$, the nonlinear activation function $\Lambda$ applied to each node, and the final readout operator $\bm{W}\in\mathbb{R}^{N_c\times d_{\mathrm{model}}}$. Note that the nonlinear activation function $\Lambda$ is applied elementwise, and is also distinct from the matrix of eigenvalues $\bm{\Lambda}$ in Eq.\,\eqref{eq:solution_basis}.

This end-to-end expression completely describes the computation performed on an individual input by a general SSM composed of a linear recurrence and nonlinear readout, and is applicable (with modification) whether the SSM dynamics is diagonalized or not. We will utilize the expression to understand computation in S4D-Lin, but note that the analysis generalizes across SSMs. The mathematical expression also makes explicit why analyzing the complete forward pass in SSMs is difficult:~even if one uses a completely standard linear recursion, which can be easily solved, the result is always passed through a nonlinear activation $\Lambda$ that aides performance on real-world classification tasks. This means that, even if the high-dimensional mappings in the linear recurrence $\bm{x}_k$ can be understood, it is very difficult to understand how the subsequent nonlinear transformation $\Lambda$ interacts with the linear transformations in $\bm{x}_k$ to accurately process real-world sequences. In this section, however, we will utilize a technique called Carleman embedding for nonlinear systems \cite{carleman1932} to understand precisely how these linear transformations interact with nonlinear activations in these SSMs.

To do this, the recurrence of the SSM layer
\begin{equation}
    \bm{x}_{k+1}=\mathcal{D}_\tau \bm{x}_k + \bar{\bm{B}} \bm{u}_k
\end{equation}
can be expressed at any future time $T$, in terms of the initial conditions and input, as
\begin{equation}
    \bm{x}_T
    =
    \mathcal{D}_\tau^{\,T} \bm{x}_{0}
    +
    \bar{\bm{B}}\sum_{j=1}^{T}
      \mathcal{D}_\tau^{\,T-j} \bm{u}_j,
    \label{eq:xT_unrolled_clean}
\end{equation}
which expresses the final state as a linear superposition of the propagated initial condition and a weighted input history. Diagonalizing the propagator $\mathcal{D}_\tau$ with the basis $\bm{F}$ leads to the modal dynamics
\begin{equation}
    \mu_i(k)
    =
    \lambda_{D,i}^{\,k}
    \mu_i(0)
    +
    \bm{\tilde{b}}_i^{T}
    \sum_{j=1}^{k}
        \lambda_{D,i}^{\,k-j}\,u_j
    \label{eq:modal_dynamics}
\end{equation}
as before (cf.\,Eq.\,\ref{eq:basic_mu_update}). Here, each eigenmode evolves independently and accumulates its own filtered input history. 

\begin{figure*}[thb]    
     \centering
     \includegraphics[width=.95\linewidth]{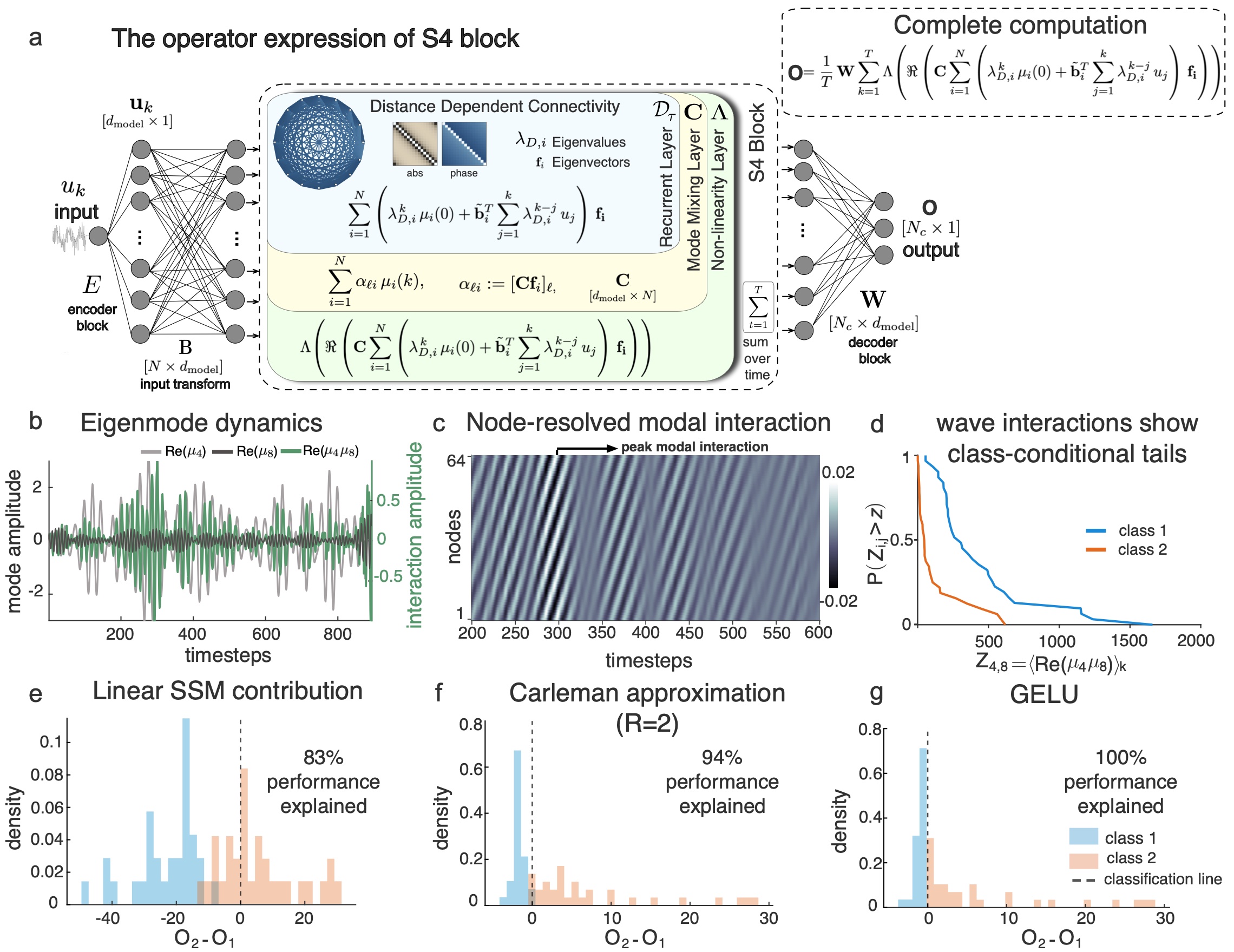}
     \caption{\textbf{Operator description of S4 performing classification on real-world input sequences.} \textbf{(a)} The input-to-output mapping in S4 admits a closed-form operator expression: the recurrent operator $\mathcal{D}_\tau$ evolves the latent state in the Fourier eigenbasis, producing modal amplitudes $\mu_i(k)$ that encode traveling-wave dynamics. The mixing matrix $\bm{C}$ forms intermediate features $\bm{y}_k$, and the nonlinear readout is expressed at the operator level via polynomial lifts of these modal coordinates. \textbf{(b)} S4 dynamics during classification of the real-world dataset SelfRegulationSCP1 \cite{bagnall2018uea}. Representative eigenmode trajectories $\mathrm{Re}(\mu_4(k))$ and $\mathrm{Re}(\mu_8(k))$ wax and wane at different sections of the input. The term $z_{4,8}(k)$ captures the nonlinear interaction between $\mu_4$ and $\mu_8$, quantifying times where the interaction between these two waves can contribute to sequence classification. \textbf{(c)} The nonlinear interaction term $z_{4,8}(k)$ in the amplitude basis illustrates how these interacting waves propagate over the network. Note how the activity pattern caused by these two interacting waves bends at different times due to time-varying phase offsets in the two waves. The time of peak wave interaction is denoted with an arrow at top. \textbf{(d)} The distributions of the interaction term $Z_{4,8}=\langle \mathrm{Re}(\mu_4\mu_8)\rangle_k$ separate two input classes (blue and red, respectively) for the SCP1 dataset. This indicates that wave-wave interactions distinguish input classes beyond the contribution of individual wave modes. \textbf{(e)} The linear SSM contribution explains 83\% of the network's classification performance, quantified in terms of the output margin $O_2-O_1$. \textbf{(f)} A second-order Carleman approximation ($R=2$), implemented via a weighted Chebyshev fit to the GELU nonlinearity (see Appendix~\ref{secA3} for details), captures quadratic modal interactions and explains 94\% of the network's performance. \textbf{(g)} The complete GELU activation function explains 100\% of the classification performance in this network, illustrating the level to which the Carleman embedding converges.}
    \label{fig:nonlinear_interactions}
\end{figure*}
Because real-world data possess much broader spectral content than the sinusoid example presented above (Fig.\,\ref{fig:dynamics}), real-world sequence inputs do not map onto single eigenmodes in a simple way, in general. The transformation $\bm{C}$ on the output of the dynamics operator mixes the wave modes, 
\begin{equation}
    s_k := \bm{C}\bm{x}_k
    =
    \sum_{i=1}^N \mu_i(k)\,c_i,
    \qquad
    c_i := \bm{C}\bm{f}_i\in\mathbb{C}^{d_{\mathrm{model}}}\,,
\end{equation}
in a manner that helps to separate features in real-world data inputs.

We can then define the observed features $\bm{y}_k$ as the real part of $s_k$
\begin{equation}
    \bm{y}_k := \Re(s_k)\in\mathbb{R}^{d_{\mathrm{model}}}.
\end{equation}
Using $\Re(z)=\frac12(z+\overline z)$, this gives
\begin{equation}
    \bm{y}_k
    =
    \frac12\sum_{i=1}^N
    \left(
        \mu_i(k)c_i
        +
        \overline{\mu_i(k)}\,\overline{c_i}
    \right).
    \label{eq:y_real_vector}
\end{equation}
Writing the $\ell$-th coordinate and defining $\alpha_{\ell i}:=[c_i]_\ell=[\bm{C}\bm{f}_i]_\ell$,
\begin{align}
    y_{k,\ell} &=
    \Re\!\left(\sum_{i=1}^N \alpha_{\ell i}\,\mu_i(k)\right), \\
    y_{k,\ell}
    &=
    \frac12\left(
        \sum_{i=1}^N \alpha_{\ell i}\,\mu_i(k)
        +
        \sum_{i=1}^N \overline{\alpha_{\ell i}}\,\overline{\mu_i(k)}
    \right).
    \label{eq:z_coord_clean}
\end{align}
The analysis to this point explains the computation in S4 up to the nonlinear activation function $\Lambda$, and we now need to analyze how $\Lambda$ transforms the linear recursion in S4. To do this, let $\Lambda:\mathbb{R}\to\mathbb{R}$ denote the GELU activation applied elementwise,
\[
\Lambda(x)=\mathrm{GELU}(x).
\]
Since $\Lambda$ is analytic, it admits an exact Carleman
representation in terms of elementwise monomials,
\begin{equation}
    \Lambda(y)
    =
    \sum_{r=0}^{\infty} a_r\, y^{\odot r},
    \label{eq:carleman_scalar}
\end{equation}
where $y^{\odot r}$ denotes elementwise powers and the coefficients
$\{a_r\}_{r\ge 0}$ are fixed by the nonlinearity.

Equivalently, defining the (infinite) Carleman lift
\begin{equation}
    z(y)
    :=
    \begin{bmatrix}
        1\\
        y\\
        y^{\otimes 2}\\
        y^{\otimes 3}\\
        \vdots
    \end{bmatrix},
\end{equation}
and diagonal-selection operators
\(
S_r\in\mathbb{R}^{d_{\mathrm{model}}\times d_{\mathrm{model}}^r}
\)
satisfying
\(
S_r(y^{\otimes r})=y^{\odot r},
\)
the nonlinearity can be written as
\begin{align}
    \Lambda(y)
    &=
    H_\infty\,z(y), \\
    H_\infty &:=
    \bigl[
        a_0 I\;\;
        a_1 S_1\;\;
        a_2 S_2\;\;
        a_3 S_3\;\;
        \cdots
    \bigr].
\end{align}

We can define a complex scalar
\begin{align} 
\label{comlex_scalar}
A_{k,\ell}
:=
\sum_{i=1}^N \alpha_{\ell i}\,\mu_i(k).
\end{align}
and use it in Eq. \eqref{eq:z_coord_clean} to rewrite the feature vectors as 
\begin{align}
y_{k,\ell}=\frac12(A_{k,\ell}+\overline{A_{k,\ell}}).
\end{align}
Then for each order $r$,
\begin{equation}
    y_{k,\ell}^{\,r}
    =
    2^{-r}
    \sum_{m=0}^{r}
    \binom{r}{m}
    A_{k,\ell}^{\,m}\,
    \overline{A_{k,\ell}}^{\,r-m}.
    \label{eq:y_power_binomial}
\end{equation}
Now, using Eq.\,\eqref{comlex_scalar} we can write the $m\text{-th}$ power of the complex scalar as
\begin{align}
A_{k,\ell}^{\,m}
&=
\sum_{i_1,\dots,i_m}
\alpha_{\ell i_1}\cdots\alpha_{\ell i_m}\,
\mu_{i_1}(k)\cdots\mu_{i_m}(k), \\
\overline{A_{k,\ell}}^{\,r-m}
&=
\sum_{j_1,\dots,j_{r-m}}
\overline{\alpha_{\ell j_1}}\cdots\overline{\alpha_{\ell j_{r-m}}}\,
\overline{\mu_{j_1}(k)}\cdots\overline{\mu_{j_{r-m}}(k)} .
\end{align}
Thus capturing the features of the data depends on nonlinear wave interactions of the form
\[
\mu_{i_1}\cdots\mu_{i_m}\,
\overline{\mu_{j_1}}\cdots\overline{\mu_{j_{r-m}}},
\]
including energy-like ($m=r-m$) and phase-sensitive terms. Thus, while Fig.\,\ref{fig:modal_energy} shows that the energy of the traveling wave modes in the recurrent layer is already enough to distinguish simple sinusoidal inputs embedded in noise, classification of real-world data requires nonlinear wave interactions that depend on their relative phase, extracting information beyond wave amplitude alone.

The explicit expression for the system's output (Eq.\,\ref{eq:operator}) then admits the exact Carleman representation
\begin{equation}
    \bm{O}
    =
    \frac{1}{T}\,
    \bm{W}\sum_{k=1}^T
    H_\infty\,z(\bm{y}_k)\,,
    \label{eq:O_carleman_exact}
\end{equation}
which can be truncated to order $R$,
\begin{widetext}
\begin{equation}
    \bm{O}
    \approx
    \frac{1}{T}
    \sum_{r=0}^{R} a_r\,2^{-r}
    \sum_{m=0}^{r}\binom{r}{m}\;
    \bm{W}\sum_{k=1}^T
    \Bigl(
        (\bm{A}\mu(k))^{\odot m}
        \odot
        (\overline{\bm{A}\mu(k)})^{\odot (r-m)}
    \Bigr),
    \label{eq:O_modal_carleman_real}
\end{equation}
\end{widetext}
where $\bm{A}\in\mathbb{C}^{d_{\mathrm{model}}\times N}$ has entries $\bm{A}_{\ell i}=\alpha_{\ell i}$. 

Equation\,\eqref{eq:O_modal_carleman_real} provides an explicit expression for how S4 maps an individual input to a specific output, providing an end-to-end description of computation in the system. Further, this expression also provides direct mechanistic insight into how computation works in S4:~the nonlinear activation function $\Lambda$ decomposes into a hierarchy of interaction orders -- where linear responses ($r = 1$) represent the waves excited by individual inputs, second-order responses ($r = 2$) capture interactions between two waves, and higher-order responses ($r \geq 3$) capture multi-wave interactions.

The advantage of the Carleman embedding is that it provides a holistic insight into the interactions in a nonlinear system \cite{amini2024carleman}, in contrast to standard techniques like Taylor expansion that approximate a nonlinear system only in a restricted neighborhood. In addition, the Carleman embedding for the S4 computation does not require an extensive expansion:~truncating the embedding only to second-order accounts for 94\% of the complete SSM's performance in classifying a real-world sequence (Fig.\,\ref{fig:nonlinear_interactions}). Truncating the Carleman embedding at higher orders brings the explained accuracy closer and closer to the complete system, but the second-order expansion already explains a substantial fraction of the accuracy achieved by the complete S4 system, while also providing succinct theoretical insight. Specifically, the first-order linear SSMs contribution -- representing the traveling waves directly evoked by the input -- accounts for 83\% of the network's classification accuracy, while the second-order contribution -- representing nonlinear interactions between two traveling waves in the SSM -- brings the explained performance up to 94\%. This provides a clear theoretical account of how S4 stores and uses past inputs in its activity state, in terms of waves and nonlinear wave interactions, in addition to opening new paths to faster and more lightweight implementations of S4 guided by this mathematical analysis.

\section*{Discussion}

A central challenge in the theory of machine learning is to understand what computations trained neural networks actually perform. While much theoretical work has provided insight into the learning dynamics of deep networks \cite{saxe2013exact} – that is, how a randomly initialized network evolves during training – theoretical analyses typically do not characterize the forward pass of the network -- where an individual input sequence drives the network to generate a specific output. The results presented here represent a first step in providing a theoretical description of the forward pass in SSMs, providing new insight into how S4 generates individual outputs through its trained connection weights.

By identifying a precise mathematical correspondence between SSMs and nonlinear oscillator networks, this work demonstrates that a large class of linear time-invariant SSMs that have achieved state-of-the-art performance on sequence processing benchmarks can be completely understood in terms of traveling waves. Complex spatiotemporal patterns – and traveling waves, in particular – have recently been of interest for computation in both state-space models and the wider class of recurrent neural networks (RNNs) \cite{heeger2019oscillatory,rusch2024oscillatory,miyato2024artificial,liboni2025image,keller2023neural,karuvally2024hidden,muzellec2025enhancing}, in addition to biological brains \cite{muller2018cortical,engel2019new}. The framework introduced here shows that these wave dynamics are not merely byproducts of the dense interconnectivity in the recurrent layer of these systems, but instead are a fundamental computational mechanism through which SSMs process sequence information.

\subsection*{Traveling waves are the computational \\ substrate of SSMs}

The correspondence established in this work reveals that diagonal SSMs like S4D-Lin are mathematically equivalent to a specific, exactly solvable nonlinear oscillator network. In the eigenbasis provided by this connection, the recurrent dynamics in S4 unfold as waves traveling around the network's one-dimensional ring topology, rather than relatively abstract transformations on a latent state space. This physical insight provides, in turn, a concrete physical meaning for S4's complex-valued parameterization: the recurrent operator $\mathcal{D}_\tau$ implements a bank of input-driven traveling wave modes, each with a specific spatial frequency.

The operator $\mathcal{D}_\tau$ performs a highly structured spectral factorization of individual input sequences. When $\mathcal{D}_\tau$ is circulant, its eigenvectors are discrete Fourier modes on a spatially organized ring topology of the network \cite{budzinski2022geometry,budzinski2023analytical}, and different input sequences selectively excite different subsets of these modes. This selective excitation produces distinct signatures of modes that store information about the recent past in a manner useful for sequence classification problems. In the relatively simple sinusoid dataset, for example, only a small set of modes~-- those whose frequencies resonate with inputs in the 15-20 Hz temporal frequency band -- serve to distinguish between input classes. Although the distribution of energy across modes is broadly similar across the different input classes, a specific set of resonant modes exhibit consistent shifts in amplitude that enable linear separation of the input classes. This spectral factorization emerges through the linear dynamics of the recurrent layer, prior to any nonlinear processing.

\subsection*{Nonlinear activation induces interactions \\ between waves}

The nonlinear activation function induces interactions between waves in the SSM's linear recurrence, amplifying the waves' ability to store information about recent inputs in a complex spatiotemporal pattern of activity in the network. Using Carleman embedding, we derived an exact operator expression for the complete forward pass of S4, revealing how the GELU activation decomposes computation into a hierarchy of interactions between wave modes. This operator expression makes explicit how S4 combines information across timescales in real-world datasets:~the readout weights determine which wave modes and nonlinear wave interactions contribute to the output classification. Truncating the Carleman embedding to first order reveals that the system stores information in the amplitude of traveling waves in the recurrent layer. Truncating the embedding to second order reveals quadratic interactions $\mu_i(k)\mu_j(k)$ that capture interaction between waves and cross-frequency couplings. Higher-order truncations correspond to increasingly complex interactions between multiple waves. The lifting mechanism of the nonlinear activation function thus transforms the wave dynamics into an interaction space whose low-order terms already capture most task-relevant structure. The mathematical analysis thus reveals that S4's computational power derives not only from its ability to maintain information about the recent past through wave dynamics, but also from the geometry of interactions between wave modes induced by the nonlinear readout.

\subsection*{Implications for understanding and \\ designing SSMs}

This mathematical framework provides several insights into why SSMs achieve state-of-the-art performance on sequence processing tasks. First, temporal information is encoded as traveling waves -- coherent spatiotemporal activity patterns, rather than abstract, high-dimensional activity patterns on a random network topology. Second, the specific eigenvalue structure in S4D-Lin ensures that, once excited, all wave modes decay at identical rates, preventing any single mode from dominating purely due to stability or an intrinsically long lifetime. This produces nearly uniform distributions of energy across modes, which can flexibly process diverse input patterns. Third, the complex-valued parameterization of S4 approximates heterogeneous time delays across connections in the recurrent layer \cite{budzinski2023analytical}, expanding the effective dimensionality of the system \cite{hart2019delayed} and enhancing its computational capacity \cite{heeger2019oscillatory,ebato2024impact,tavakoli2024boosting,marzen2024time,muzellec2025enhancing}.

These physical insights also extend beyond simple traveling wave patterns in the recurrent layer. When connectivity or phase-lag parameters are altered, a specific set of modes could be designed to have similar decay rates, resulting in more complex interference patterns than occur in S4D-Lin by default. This could produce longer-lived transient spatiotemporal structures that could, in turn, be useful for capturing longer-time input dependencies, and could further improve SSMs' capabilities relative to transformers. The mathematical framework introduced in this work can thus suggest specific, mathematically guided design principles for future SSM architectures. In addition, the analysis presented here could be directly extended to architectures with time-varying recurrences, such as Mamba \cite{gu2023mamba}, in future work.

\subsection*{A new level of mathematical interpretability \\ for SSMs}

Current methods for explaining machine learning architectures for sequence processing or language modeling require analyzing outputs {\it post-hoc} in order to trace model inputs to outputs. This process is largely done manually, and is also data- and method-dependent, meaing that different choices of input datasets or analysis parameters yield different explanations of the model. Thus, while approaches like mechanistic interpretability can provide insight into how systems like large language models process inputs and generate outputs \cite{ nanda2023progress, elhage2021mathematical}, these methods remain fundamentally empirical, observing model output rather than deriving first-principles insight from the structure of the model itself.

In contrast, the analysis introduced here opens the door to understanding the input-output mapping at a mathematical level, enabling prediction of outputs {\it a priori} without necessarily needing even to run the model. By deriving an explicit operator expression (Eq.\,\ref{eq:O_modal_carleman_real}) that relates arbitrary inputs to model ouputs in a classification task, we establish a mathematical framework where the forward pass becomes analytically tractable. This represents a qualitative shift from needing to interpret model output {\it post-hoc} to prospective understanding:~rather than empirically probing a trained model to understand its computational strategy, we can derive how specific input features propagate through the recurrent dynamics, undergo nonlinear transformation, and lead to the model output.

This mathematical interpretability can lead to several clear benefits, including the potential for controllable language models in future work. If the mapping from input sequences to model output can be understood in terms of explicit mathematical expressions, then one can potentially design eigenvector-based strategies to control model output in real time. This level of control is fundamentally unavailable when methods for interpretability rely solely on empirical observation rather than mathematical expressions for computation. Such controllable language models, based on mathematical expressions, would enable a new level of guarantees on model output and safety that are critical to deployment in sensitive applications.

\subsection*{Broader implications}

This work demonstrates that state-of-the-art machine learning architectures can be precisely understood through explicit mathematical expressions. The operator expression for S4 provides the first end-to-end characterization of computation in a modern sequence processing architecture, revealing how traveling-wave recurrent dynamics and nonlinear modal interactions combine to produce strong sequence classification performance. By bridging nonlinear dynamical systems, spectral graph theory, and modern deep learning architectures, this framework opens new directions for understanding sequence processing and language modeling at a mathematical level.

\section*{Code availability}

An open-source version of the code and data used in this study will be made available at \href{https://mullerlab.github.io}{\textcolor{Cerulean}{https://mullerlab.github.io}}.

\begin{acknowledgements}
This work was supported by BrainsCAN at Western University through the Canada First Research Excellence Fund (CFREF), the NSF through a NeuroNex award (\#2015276), the Natural Sciences and Engineering Research Council of Canada (NSERC) grant R0370A01, Compute Ontario (computeontario.ca), Digital Research Alliance of Canada (alliancecan.ca), the Air Force Office of Scientific Research (FA9550-24-1-0028), and NIH Grants U01-NS131914 and R01-EY028723.
\end{acknowledgements}

%


\clearpage

\section{Appendix}

\subsection{Closed-form diagonalization of circulant operators}\label{secA1}

Let $\bm{C}\in\mathbb{C}^{N\times N}$ be a circulant matrix generated by the vector
\[
\bm{c}=(c_1,c_2,\dots,c_N),
\]
such that each row of $\bm{C}$ is obtained by a cyclic
shift of $\bm{c}$. By the Circulant Diagonalization Theorem, $\bm{C}$ is diagonalized
by the unitary discrete Fourier transform matrix $\bm{F}$ with entries
\begin{equation}
\begin{split}
f_{js}=\frac{1}{\sqrt{N}}
\exp\!\left[-\frac{2\pi \i}{N}(s-1)(j-1)\right], \\
j,s=1,\dots,N .
\end{split}
\end{equation}
The corresponding eigenvalues admit a closed-form expression in terms of the
generating vector $\bm{c}$,
\begin{equation}
\begin{split}
\lambda_s
=
\sum_{j=1}^{N} c_j
\exp\!\left[-\frac{2\pi \i}{N}(s-1)(j-1)\right], \\ s=1,\dots,N ,
\end{split}
\end{equation}
which is simply the discrete Fourier transform of $\bm{c}$. As a result, the
circulant operator can be written as
\begin{equation}
\bm{C}=\bm{F}\,\mathrm{diag}(\lambda_1,\dots,\lambda_N)\,\bm{F}^\dagger.
\end{equation}
Applying the change of basis $\bm{F}^\dagger$ to the input-driven recurrence in
Eq.~\eqref{ko20input} therefore yields a diagonal state update in the Fourier
eigenbasis, in which each mode evolves independently.

\subsection{Example of several diagonal SSMs in this mathematical framework}\label{secA2}
\begin{figure}[t!]

    \centering
    \includegraphics[width=0.9\linewidth]{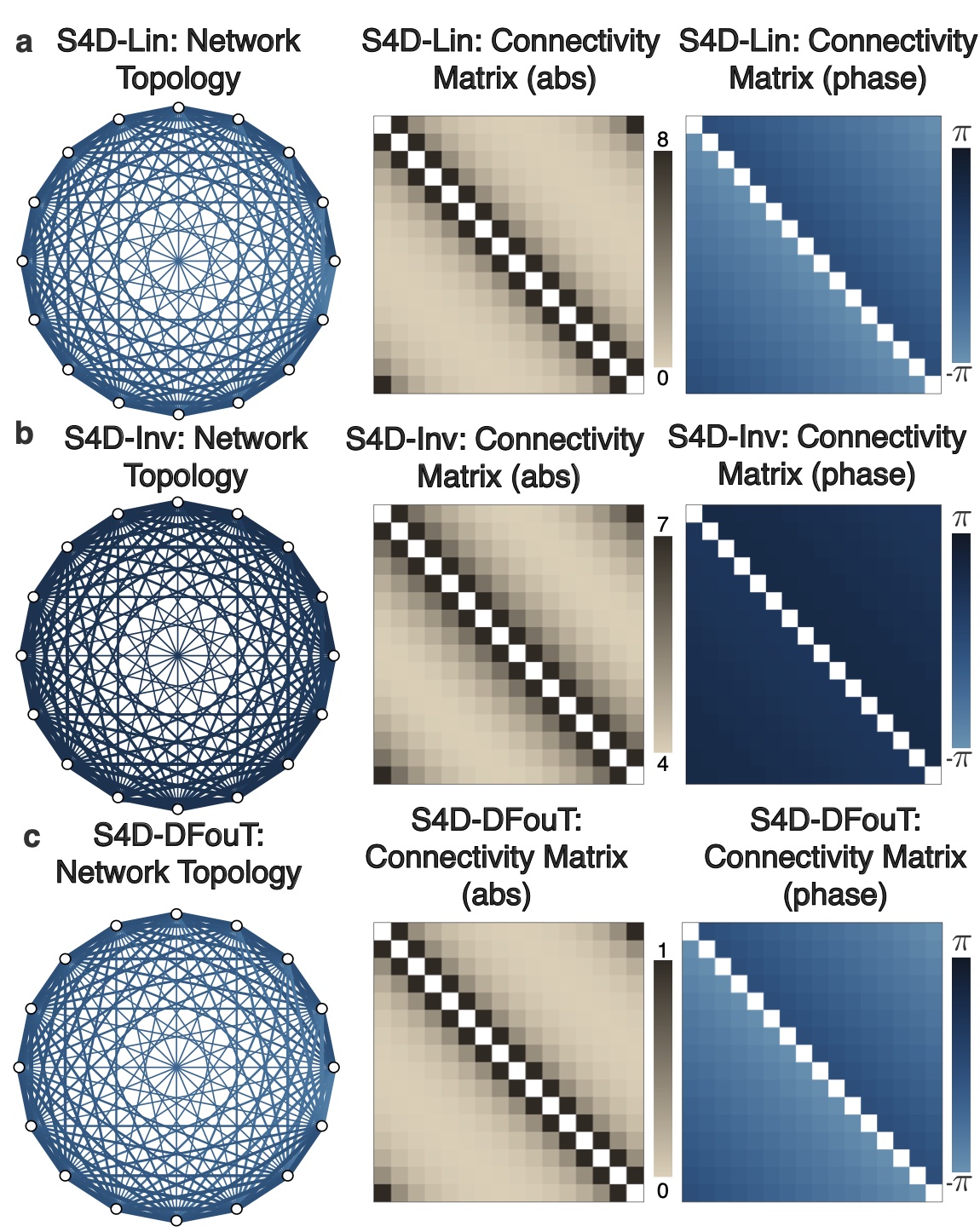}
    \caption{\textbf{Topology and induced connectivity structure of the S4D operators.} Each row shows the network representation and connectivity matrices induced by the eigenvalue spectrum of the diagonal recurrence operator. \textbf{(a)} \textbf{S4D-Lin}: $\lambda_n = -\tfrac{1}{2} + \i\pi n$, yielding identical decay rates and linearly spaced imaginary parts and a circulant, Fourier-diagonalizable propagator. \textbf{(b)} \textbf{S4D-Inv}: $\lambda_n = -\tfrac{1}{2} + \i\,\tfrac{N}{\pi}\!\left(\tfrac{N}{2n+1}-1\right)$, inverting the spectral ordering while preserving stability. \textbf{(c)} \textbf{S4D-FouTD}: $\lambda_n = -\tfrac{1}{2} + \i\,\tfrac{2\pi n}{N}$, corresponding to a uniform Fourier frequency grid; differences in the displayed connectivity arise solely from this eigenvalue parameterization.}
    \label{fig:S4D_topologies}
\end{figure}

Figure \ref{fig:S4D_topologies} compares several diagonal SSMs by visualizing the networks that result from translating their recurrence operators into the oscillator network context, using the expression $\bm{K} = \bm{F} \bm{D} \bm{F}^\dagger$ (as in Fig.\,\ref{fig:dynamics}). The resulting network adjacency matrix for S4D-Lin has connection strengths that decay with distance between nodes, and phase offsets that change linearly with distance (Fig.\,\ref{fig:S4D_topologies}a). The network adjacency matrix resulting from S4D-Inv has connection strengths that decrease similarly with distance between nodes (Fig.\,\ref{fig:S4D_topologies}b, middle panel), but these strengths decrease more slowly than for S4D-Lin. Phase offsets in the resulting network are nearly constant (Fig.\,\ref{fig:S4D_topologies}b, right panel). The network adjacency matrix for S4D-DFouT shows similar decay in connection strength and change in phase offset to S4D-Lin (Fig.\,\ref{fig:S4D_topologies}c). In all cases, the recurrent evolution admits an exact modal decomposition, and differences between parameterizations arise from eigenvalue allocation and readout structure rather than from changes in the linear propagator itself; consequently, each variant is amenable to the same modal and operator-level analysis developed in the main text.

\subsection{Chebyshev Approximation of Carleman Embedding}\label{secA3}

For Fig.~\ref{fig:nonlinear_interactions}, we implement a truncated polynomial approximation of the point-wise nonlinearity using Chebyshev polynomials. Chebyshev polynomials of the first kind are defined by
\begin{equation*}
    T_0(x)=1,\,\,\, T_1(x)=x,\,\,\, T_{k+1}(x)=2xT_k(x)-T_{k-1}(x)
\end{equation*}
and span the same degree-\(K\) polynomial space as monomials, while offering improved numerical stability.

Given readout activations \(y\), we rescale and clip them to the interval \([-1,1]\) and construct the lifted features
\begin{equation*}
    z(y)=\bigl[T_0(y),\,T_1(y),\,\dots,\,T_K(y)\bigr].
\end{equation*}
The nonlinearity is approximated by a truncated Chebyshev series fitted via weighted least squares, giving a polynomial (Carleman) feature map applied at the readout.

\subsection{Training S4D on a sequence processing dataset}

To demonstrate how S4D processes real-world sequence inputs, we trained a single-layer S4D architecture on the SelfRegulationSCP1 (SCP1) classification task \cite{bagnall2018uea}. The SCP1 dataset is a multivariate, long-horizon time-series dataset collected during a self-regulation paradigm, where the goal is to infer cognitive control states from temporally extended neural signals. For the purpose of probing the contribution of the nonlinear (higher-order) interactions to class separation, we perform an analysis on the test set in which misclassified samples are excluded. This is done to isolate the mechanisms underlying successful discrimination by restricting focus to only correctly classified examples.

For the mathematical analysis presented in the paper, the eigenvalues (or the propagator $\mathcal{D}_{\tau}$) are kept fixed while the readout matrix $\bm{W}$ and the mode mixing matrix $\bm{C}$ are trained. The discretization step $\tau$ (equivalently $\Delta t$) is fixed to $0.01$ in all experiments. Unless otherwise stated, the state dimension and model width are set to $N = 64$ and $d_{\text{model}} = 64$, respectively. Model parameters are optimized using the AdamW optimizer with standard hyperparameter settings.

\end{document}